\documentclass[letterpaper]{article} % DO NOT CHANGE THIS
\usepackage{aaai24}  % DO NOT CHANGE THIS
\usepackage{times}  % DO NOT CHANGE THIS
\usepackage{helvet}  % DO NOT CHANGE THIS
\usepackage{courier}  % DO NOT CHANGE THIS
\usepackage[hyphens]{url}  % DO NOT CHANGE THIS
\usepackage{graphicx} % DO NOT CHANGE THIS
\usepackage{natbib}  % DO NOT CHANGE THIS AND DO NOT ADD ANY OPTIONS TO IT
\usepackage{caption} % DO NOT CHANGE THIS AND DO NOT ADD ANY OPTIONS TO IT
\usepackage{algorithm}
\usepackage{algorithmic}

\usepackage{newfloat}
\usepackage{listings}
\DeclareCaptionStyle{ruled}{labelfont=normalfont,labelsep=colon,strut=off} % DO NOT CHANGE THIS
\floatstyle{ruled}
\newfloat{listing}{tb}{lst}{}
\floatname{listing}{Listing}
\usepackage{amsmath}
\usepackage{amsthm}
\usepackage{amsfonts}
\usepackage{multirow}
\usepackage{color}
\usepackage{subcaption}
\usepackage{booktabs}

\title{Point Cloud Part Editing: Segmentation, Generation, Assembly, and Selection}
\author{
    Kaiyi Zhang\textsuperscript{\rm 1},
    Yang Chen\textsuperscript{\rm 1},
    Ximing Yang\textsuperscript{\rm 1},
    Weizhong Zhang\textsuperscript{\rm 1,2},
    Cheng Jin\textsuperscript{\rm 1,2}
}
\affiliations{
    \textsuperscript{\rm 1}School of Computer Science, Fudan University, Shanghai, China\\
    \textsuperscript{\rm 2}Innovation Center of Calligraphy and Painting Creation Technology, MCT, China\\
    \{zhangky20, chen\_yang19, xmyang19, weizhongzhang, jc\}@fudan.edu.cn
}

\usepackage{bibentry}
\begin{document}

\maketitle

\begin{abstract}
  Ideal part editing should guarantee the diversity of edited parts, the fidelity to the remaining parts, and the quality of the results. However, previous methods do not disentangle each part completely, which means the edited parts will affect the others, resulting in poor diversity and fidelity. In addition, some methods lack constraints between parts, which need manual selections of edited results to ensure quality. Therefore, we propose a four-stage process for point cloud part editing: \textbf{S}egmentation, \textbf{G}eneration, \textbf{A}ssembly, and \textbf{S}election. Based on this process, we introduce \textbf{SGAS}, a model for part editing that employs two strategies: feature disentanglement and constraint. By independently fitting part-level feature distributions, we realize the feature disentanglement. By explicitly modeling the transformation from object-level distribution to part-level distributions, we realize the feature constraint. Considerable experiments on different datasets demonstrate the efficiency and effectiveness of SGAS on point cloud part editing. In addition, SGAS can be pruned to realize unsupervised part-aware point cloud generation and achieves state-of-the-art results.
\end{abstract}

\section{Introduction}

In the context of 3D object modeling, parts are considered the fundamental units. Recently, part-based methods~\citep{Wang_TOG_2018,mo2019structurenet,Li_Niu_Xu_2020,jones2020shapeAssembly,Gal_2021_ICCV,Li_Liu_Walder_2022} have become more and more prevailing. These methods typically involve obtaining different parts first and then assembling them. Although many works have explored procedural content generation~\citep{Liu_2021_NCA}, which is often used to make material maps and game maps, the rapid development of game scenes still relies heavily on the generation of 3D objects. Part-based methods enable part editing, which involves replacing some parts of an object to create a new one, thereby enhancing the diversity of 3D modeling.

\begin{figure*}[htb]
    \centering
    \includegraphics[width=\textwidth]{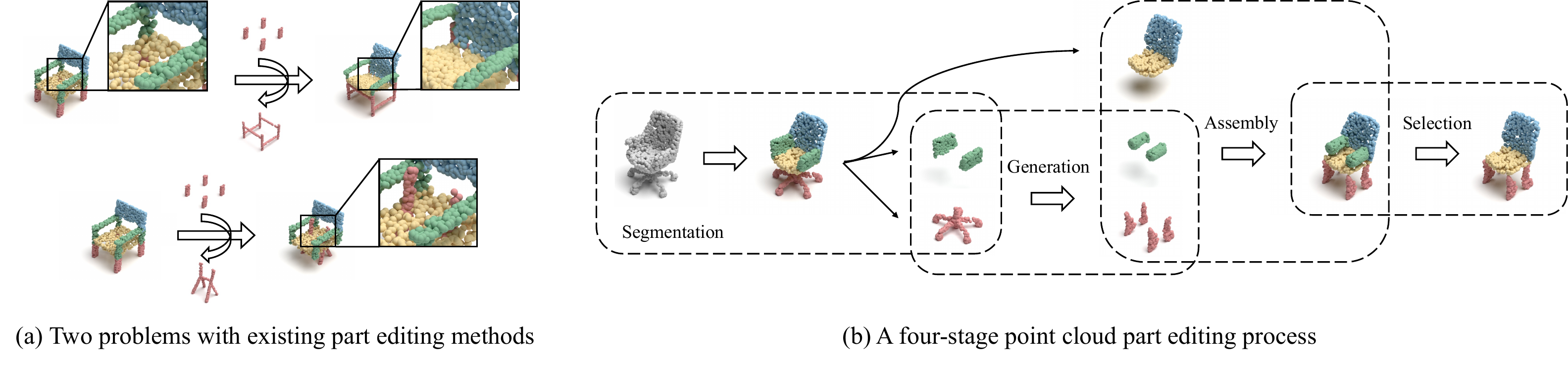}
    \caption{(a) Top: w/o disentanglement between parts. When changing the chair base, the other parts such as the chair back, arm, and seat will also change. Bottom: w/o constraints between parts. The changed chair base is not only poorly assembled but also does not match other parts, resulting in the generation of an unreasonable object. (b) It includes four stages: segmentation, generation, assembly, and selection, which can guarantee fidelity, diversity, and quality of the edited results respectively.}
    \label{fig:intro}
\end{figure*}

Ideal part editing should make edited parts diverse while keeping unedited parts unchanged to form a reasonable object. These correspond to three important properties of the edited results: diversity, fidelity, and quality. However, when previous part-based methods are applied to part editing, they have two problems. As shown in Figure~\ref{fig:intro}(a), on the one hand, methods such as MRGAN~\citep{Gal_2021_ICCV} and SP-GAN~\citep{li2021spgan} do not realize radical disentanglement between parts. Therefore, when some parts are modified, other parts will also change, which means they do not guarantee the fidelity to the remaining parts. Similarly, multimodal shape completion methods such as MSC-cGAN~\citep{wu2020multimodal}, which can be regarded as a subset of part editing, also do not disentangle parts. This not only changes the input parts but also makes the completion results less diverse. Someone may argue that the adjacent parts may change to accommodate the edited parts, but this will not affect the requirement of radical disentanglement, since all changed parts can be regarded as edited parts. On the other hand, some methods~\citep{Schor_2019_ICCV,Li_Niu_Xu_2020,Li_Liu_Walder_2022} do not implement constraints between parts effectively. This may result in poor part assembly and mismatched parts to be used in the formation of an object. Although these methods attempt to achieve assembly by moving parts, the changed parts still need a manual selection to ensure high-quality edited results that lead to a reasonable object.

To address these issues, as shown in Figure~\ref{fig:intro}(b), we first propose a four-stage process for point cloud part editing: segmentation, generation, assembly, and selection. For the three properties of edited results, segmentation can guarantee the fidelity to the remaining parts by isolating the parts; generation can guarantee the diversity of edited parts by exploring different variations of the parts; assembly and selection can guarantee the quality of the results by choosing the most appropriate edited parts and assembling them in a coherent manner. Based on this process, we introduce a model SGAS for part editing that employs two strategies: feature disentanglement and constraint. We first use unsupervised shape co-segmentation methods~\citep{Chen_2019_ICCV,Zhu_2020_CVPR,Zhang_Yang_Wu_Jin_2022} or manual segmentation to obtain Ground Truth parts. Then we pre-train several autoencoders at the part level. Finally, by adversarially supervising part-level feature transformations, we realize the feature disentanglement during generation. Since each part is generated separately, this not only ensures the diversity of edited parts but also the fidelity to the remaining parts. In addition, we make the distribution of each part transformed from the same Gaussian distribution and adversarially supervise the generations of all parts simultaneously to ensure that edited parts can assemble well and form a reasonable object. This strategy is called feature constraint. It guarantees the quality of the edited results. By adding a part selection module to the final output part features, which allows SGAS to autonomously choose which parts do not need to be output, the quality of the edited results can be further improved.

Our main contributions are the following:
\begin{itemize}
    \item We propose a novel point cloud part editing process. It inlcudes four stages: segmentation, generation, assembly, and selection.
    \item Based on the proposed process, we introduce SGAS, a model for part editing that employs two strategies of feature disentanglement and constraint. Experiments show that SGAS achieves excellent quantitative and qualitative part editing results.
    \item A new diversity metric of edited results: Total Mutual Difference Surface (TMDS).
    \item SGAS can be pruned to realize unsupervised part-aware point cloud generation and achieves state-of-the-art results on the ShapeNet-Partseg dataset.
\end{itemize}

\section{Related Work}

\subsection{Part-based Shape Generation}

\subsubsection{Unsupervised Part-aware Point Cloud Generation}

The fine-grained improvement of generative results can be achieved through local generation. Therefore, many methods attempt to explore the generation of multiple parts and combine them into a final shape. Since part-level ground truth data is often unavailable, these methods typically involve unsupervised segmentation of parts. For example, TreeGAN~\citep{Shu_2019_ICCV} first designs the generation of the point cloud as a tree growth process and then combines the various parts at the leaf nodes. To achieve controllable point cloud generation, SP-GAN~\citep{li2021spgan} is proposed. Similar to FoldingNet~\citep{Yang_2018_CVPR}, SP-GAN transforms a sphere in 3D space into a target point cloud, where different parts of the 3D sphere correspond to different parts of the target point cloud. MRGAN~\citep{Gal_2021_ICCV} explicitly realizes part disentanglement by using multiple branches of tree-structured graph convolution layers. Instead of supervising each part respectively, it conducts overall supervision after assembling all the parts. Considering that the parts of MRGAN lack semantic meaning, \citet{Li_Liu_Walder_2022} propose EditVAE, which can achieve semantics-aware point cloud generation. Each branch of EditVAE generates not only parts but also additional part offsets and primitives for auxiliary supervision. In addition, \citep{ngn2020LPMNetLP,Postels_2021_3DV,Li_2022_TVCG,cheng2022autoregressive} also play important roles in promoting part-aware point cloud generation.

\subsubsection{Assembly-based Shape Generation}

Many datasets, such as ShapeNet-Partseg~\citep{Yi16} and PartNet~\citep{Mo_2019_CVPR}, provide part-level semantics. Therefore, many works explore shape generation by assembling parts. Specifically, these works can be roughly categorized into three groups: (1) assemble without generation~\citep{Schor_2019_ICCV,Dubrovina_2019_ICCV,yin2020coalesce,Hui_2022_CVPR,wu2023attention}. These methods only reconstruct the parts and achieve the diversity of results by assembling different parts. For example, CompoNet~\citep{Schor_2019_ICCV} synthesizes "unseen" but reasonable point clouds by varying both the parts and their compositions. \citet{Dubrovina_2019_ICCV} propose a semantic-part-aware embedding space to realize shape composition and decomposition. PartAttention~\citep{wu2023attention} uses a part-wise attention framework to achieve affine transformation of the decoded parts. (2) assemble after generation~\citep{li_sig17,Wang_TOG_2018,Wu_TOG_2019,Li_Niu_Xu_2020}. In contrast to the first category, the parts used for assembling are generated from a Gaussian distribution. These methods mainly focus on designing part offset networks to efficiently assemble parts. For example, G2LGAN~\citep{Wang_TOG_2018} uses global and local GANs to supervise the correlation between the parts and the quality of each part respectively. It also adds a Part Refiner to optimize the generated results, such as removing outliers and completing missing regions. PAGENet~\citep{Li_Niu_Xu_2020} generates parts using part-level VAEs and designs a Part Assembler to translate parts based on some anchor parts. (3) assemble while generating~\citep{Zou_2017_ICCV,mo2019structurenet,Mo_2020_CVPR,Wu_2020_CVPR,jones2020shapeAssembly,wang2022shape,Zhuang_2022}. This type of method does not assemble the parts after generating all of them, but rather assembles them progressively during generation. For example, 3D-PRNN~\citep{Zou_2017_ICCV} proposes a generative recurrent neural network that synthesizes multiple plausible shapes step-by-step based on primitives. This progressive process preserves long-range structural coherence. PQ-Net~\citep{Wu_2020_CVPR} adopts RNN structure and learns 3D shape representations as a sequential part assembly. ShapeAssembly~\citep{jones2020shapeAssembly} achieves 3D shape structure synthesis by generating domain-specific language programs. The transformation of the statements in these programs enables ShapeAssembly to control the generated results.

\subsection{Multimodal Shape Completion}

Shapes with missing semantics can lead to a variety of completion results. For example, \citet{wu2020multimodal} propose MSC-cGAN. Based on pcl2pcl~\citep{chen2020pcl2pcl}, it adds an additional Gaussian distribution during the transformation from partial to complete point cloud features and an encoder to guarantee completion fidelity to the input partial. Different samples on Gaussian distribution correspond to different completion results. \citet{Zhou_2021_ICCV} introduce PVD, a unified probabilistic formulation, to achieve multimodal shape completion by progressively removing noise from the samples. AutoSDF~\citep{Mittal_2022_CVPR} utilizes a transformer-based autoregressive model to generate patch embeddings extracted independently by VQ-VAE~\citep{NIPS2017_7a98af17} step-by-step. The idea of ShapeFormer~\citep{Yan_2022_CVPR} is similar to AutoSDF. The difference is that AutoSDF embeds the whole 3D space while ShapeFormer introduces a compact 3D representation VQDIF that embeds only the space occupied by 3D shapes, making it more efficient. There are also many other works~\citep{Arora_2022_CVPR,zhang2021gca,Zhao_2021_MM,jiang2022probabilistic,Cheng_2023_CVPR} exploring multimodal shape completion.

\begin{figure*}[htb]
    \centering
    \includegraphics[width=0.8\textwidth]{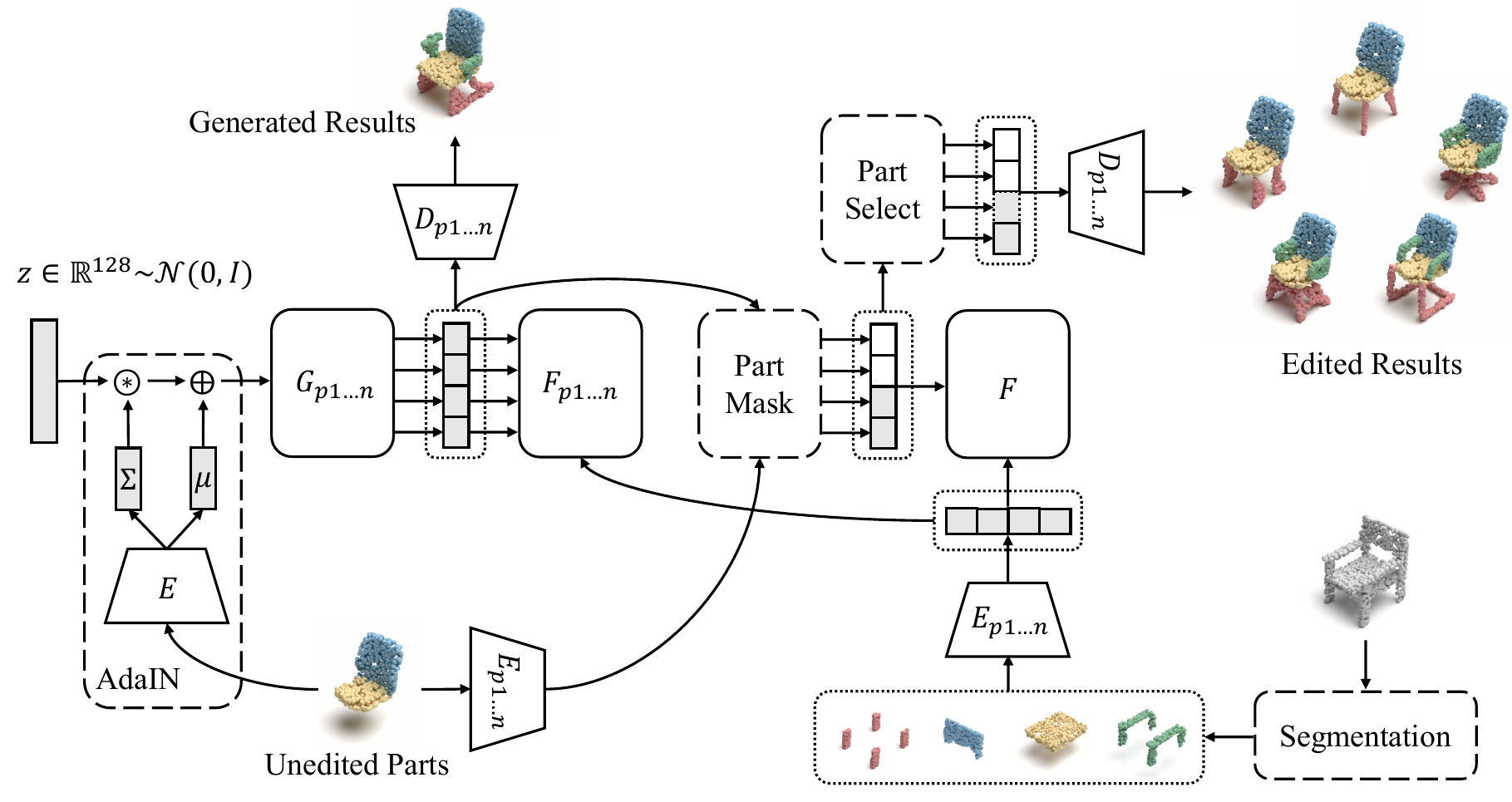}
    \caption{The architecture of SGAS. The inputs are Gaussian noise and unedited parts. The outputs are diverse generated or edited results. SGAS obtains Ground Truth parts through segmentation and uses them to pre-train part-level autoencoders, which convert point clouds into features. By incorporating the style of unedited parts into the Gaussian distribution using an AdaIN layer and performing part-level GAN supervision, SGAS can generate new parts. To constrain each part to form a reasonable object, SGAS applies part masking and uses a global discriminator. Finally, SGAS performs part selection on each part feature, allowing the model to autonomously choose which parts do not need to be output.}
    \label{fig:method}
\end{figure*}

\section{Method}

In this section, we first describe the architecture of SGAS according to the proposed four-stage point cloud part editing process, and then give the loss functions of SGAS.

\subsection{Segmentation}

To achieve the disentanglement between parts which can guarantee fidelity in part editing, SGAS is designed to have multiple branches. Each branch generates a part and requires a Ground Truth part for supervision. In our opinion, parts can be semantic parts, such as a chair back, seat, and base, and can also be local areas of a shape's surface. The former can directly use some datasets~\citep{Yi16,Mo_2019_CVPR} with part semantic labels to obtain Ground Truth parts, while the latter need some unsupervised shape co-segmentation methods~\citep{Chen_2019_ICCV,Zhu_2020_CVPR,Zhang_Yang_Wu_Jin_2022} to obtain Ground Truth parts. We follow the idea of l-GAN~\citep{achlioptas2017latent_pc}, which demonstrates that generating on features is better than directly generating on point clouds. Therefore, using the segmented Ground Truth parts, we pre-train an autoencoder for each semantic part to convert point clouds into features. The encoder is the same as PointNet~\citep{Qi_2017_CVPR} encoder. The decoder uses a fully connected network. Earth Mover’s Distance(EMD)~\citep{Fan_2017_CVPR} is used to supervise the training of these autoencoders. As shown in Figure~\ref{fig:method}, the trained encoders $E_{pi}, i=1...n$ and decoders $D_{pi}, i=1...n$ are used to build SGAS. They do not update parameters during SGAS training.

\subsection{Generation}

The input of SGAS includes not only Gaussian noise but also unedited parts. The purpose is to realize that the style of generated parts matches that of unedited parts, thereby ensuring the quality of the final edited results. We use AdaIN Layer~\citep{Huang_2017_ICCV} to integrate the unedited parts into the Gaussian distribution. Specifically, a PointNet encoder $E$ is used to encode the unedited parts to the mean $\mu$ and standard deviation $\Sigma$ of a Gaussian distribution. The $\mu$ and $\Sigma$ are then applied to the standard Gaussian distribution to obtain a new Gaussian distribution $\mathcal{N}(\mu, \Sigma^2)$. Based on this new distribution, we design several part-level GANs to generate parts. As shown in Figure~\ref{fig:method}, a Gaussian noise $z \in \mathbb{R}^{128} \sim \mathcal{N}(\mu, \Sigma^2)$ is transformed by generators $G_{pi}, i=1...n$ into part latent features to realize feature disentanglement. The generator uses a 3-layer fully connected network $(256, 512, 128)$. The dimension of the part latent feature is 128. Part features are then sent to discriminators $F_{pi}, i=1...n$ to distinguish real parts and generated parts. The discriminator uses a 3-layer fully connected network $(256, 512, 1)$. These part-level discriminators ensure the quality of each part.

\subsection{Assembly}

Since the branches used for part generation are independent, the generated parts may not be assembled into a reasonable object. To solve this problem, as shown in Figure~\ref{fig:method}, we add a global discriminator $F$ in SGAS to supervise all generated parts simultaneously, which realizes feature constraint. The discriminator uses a 3-layer fully connected network $(256, 512, 1)$. Compared to methods~\citep{Schor_2019_ICCV,Dubrovina_2019_ICCV,yin2020coalesce,Li_Niu_Xu_2020} that use affine transformation to realize part assembly, our global discriminator can further ensure matching between the generated parts. Since some parts of the target point cloud already exist, we use Part Mask to replace some generated parts. Specifically, we first use pre-trained encoders $E_{pi}, i=1...n$ to obtain the part latent features of unedited parts. These part latent features are then used to replace the corresponding generated part features. Finally, the replaced features are sent to discriminator $F$.

\subsection{Selection}

In a real scene, an object does not necessarily contain all semantic parts. For example, a chair without arms and a lamp without a holder. In order to realize this, we perform Part Select on the part features processed after Part Mask. Part Select uses a threshold $\tau$ to filter the parts that SGAS thinks do not need to be output. It does not need training. Specifically, since some parts might not exist in the Ground Truth point clouds, we set the latent features corresponding to these parts to zero. Therefore, the trained SGAS can automatically determine whether a part needs to be output. It forces the features of parts that do not need to be output as close to zero as possible. In this way, By not decoding the parts whose features are within the threshold $\tau$, we realize part selection in the output point clouds. The filter conditions for Part Select are given as:
\begin{equation}
    \left|\frac{1}{n} \sum_{i=1}^{n} G_{pi}(z)\right| \leq \tau
\end{equation}
where $|\cdot|$ represents absolute value, $n$ is the number of parts.

\subsection{Loss Functions}

We adopt the loss function introduced in Wasserstein GAN~\citep{pmlr-v70-arjovsky17a} with gradient penalty~\citep{NIPS2017_892c3b1c}. Network $E$, $G_{pi}, i=1...n$, $F_{pi}, i=1...n$, and $F$ need training. The losses are given as: 
\begin{equation}
\begin{gathered}
    \mathcal{L}_G = -\alpha*\frac{1}{n}\sum_{i=1}^n \mathbb{E}_{z \sim Z}[F_{pi}(G_{pi}(z))] \\
    - \beta*\mathbb{E}_{z \sim Z}[F(\bigcup_{i=1}^n G_{pi}(z))]
\end{gathered}
\end{equation}
\begin{equation}
\begin{gathered}
    \mathcal{L}_{F_p} = \frac{1}{n}\sum_{i=1}^n (\mathbb{E}_{z \sim Z}[F_{pi}(G_{pi}(z))] 
    - \mathbb{E}_{x_i \sim R_{xi}}[F_{pi}(x_i)] \\
    + \lambda_{gp} \mathbb{E}_{\hat{x_i}}[(\lVert \nabla_{\hat{x_i}} F_{pi}(\hat{x_i}) \rVert_2 - 1)^2])
\end{gathered}
\end{equation}
\begin{equation}
\begin{gathered}
    \mathcal{L}_F = \mathbb{E}_{z \sim Z, x_i \sim P_{xi}}[F(\bigcup_{i=1}^n M_i G_{pi}(z)) 
    + (1-M_i) E_{pi}(x_i)] \\
    - \mathbb{E}_{x_i \sim R_{xi}}[F(\bigcup_{i=1}^n x_i)] 
    + \lambda_{gp} \mathbb{E}_{\hat{x_i}}[(\lVert \nabla_{\hat{x_i}} F(\bigcup_{i=1}^n \hat{x_i}) \rVert_2 - 1)^2]
\end{gathered}
\end{equation}
where $\mathcal{L}_G$, $\mathcal{L}_{F_p}$, and $\mathcal{L}_F$ represent the loss functions of the part-level generator, the part-level discriminator, and the global discriminator respectively. $\alpha$ and $\beta$ are hyperparameters that control the proportion of the part-level to the global GAN. $N$ is the number of parts. $x_i, i=1...n$ are parts. The formulas after $\lambda_{gp}$ are the gradient penalty terms proposed by \citet{NIPS2017_892c3b1c}. $Z=\mathcal{N}(\mu, \Sigma^2)$, where $\mu$ and $\Sigma$ are from encoder $E$. $M$ is determined by the unedited parts.

\section{Experiments}

\subsection{Datasets and Implementation Details}

We evaluate SGAS on PartNet~\citep{Mo_2019_CVPR} dataset. By merging fine-grained semantic labels and removing some special objects, we create a new dataset called PartNet.v0.Merged for point cloud part editing. Following previous works~\citep{Gal_2021_ICCV,Li_Liu_Walder_2022}, we perform unsupervised part-aware point cloud generation on ShapeNet-Partseg~\citep{Yi16} dataset. We do not use semantic labels in the ShapeNet-Partseg dataset.

\begin{table}[htb]
    \small
    \centering
    \caption{Diversity part editing performance. MMD and TMD measure the quality and diversity respectively.}
    \label{tab:editings_metrics}
    \begin{tabular}{c c *4c}
        \toprule
        & Model & Chair & Lamp & Table & Average \\
        \midrule
        \multirow{2}{*}{MMD $\downarrow$} & MSC-cGAN & 1.62 & 3.41 & 1.39 & 2.14 \\
        & SGAS & \textbf{1.33} & \textbf{2.26} & \textbf{1.06} & \textbf{1.55} \\
        \midrule
        \multirow{2}{*}{TMD $\uparrow$} & MSC-cGAN & \textbf{5.45} & 3.94 & 5.14 & 4.84 \\
        & SGAS & 4.36 & \textbf{4.48} & \textbf{8.04} & \textbf{5.63} \\
        \bottomrule
    \end{tabular}
\end{table}

\begin{figure}[htb]
    \centering
    \includegraphics[width=\columnwidth]{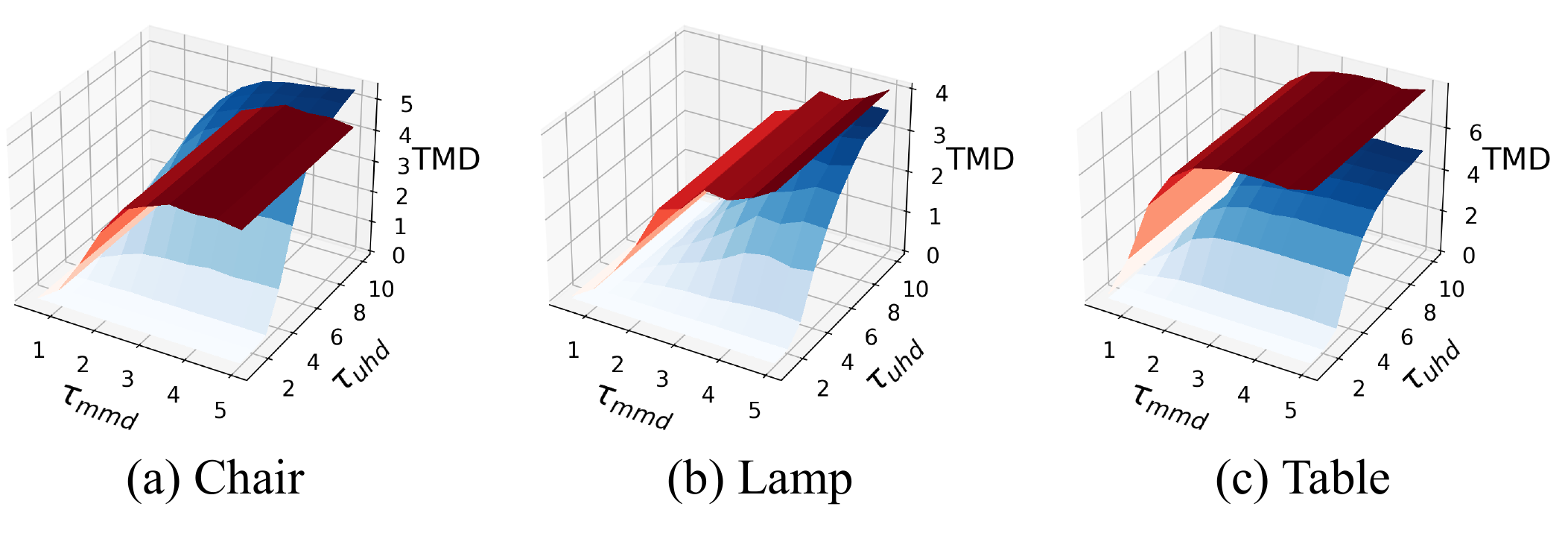}
    \caption{Performance on the new metric Total Mutual Difference Surface (TMDS). Red represents SGAS; blue represents MSC-cGAN. The smaller the thresholds of MMD and UHD are, the more referential the calculated TMD is.}
    \label{fig:tmds}
\end{figure}

Adam optimizers are used for SGAS with a learning rate of $\alpha=0.0005$, coefficients $\beta_1 = 0.5$ and $\beta_2 = 0.99$. All the experiments are performed on a single NVIDIA TITAN Xp for 2000 epochs with a batch size of 200. In loss functions, $\alpha$ and $\beta$ are set to 1 and 1. $\lambda_{gp}$ is set to 10. The threshold in Part Select is set to 0.5. We update the discriminator 5 times for each update of the generator. Each shape has 2048 points while each part has $\left\lfloor\frac{2048}{n}\right\rfloor$ points. $n$ is the number of parts. During training, the input unedited parts for SGAS are obtained by randomly removing 1 to $n-1$ parts of objects in PartNet.v0.Merged dataset.

\subsection{Point Cloud Part Editing}

Diversity part editing is a commonly used operation in point cloud part editing that involves generating some parts multiple times to obtain various results. Previous part editing methods~\citep{Li_Niu_Xu_2020,Gal_2021_ICCV,li2021spgan,Li_Liu_Walder_2022} lack related evaluation metrics. Since diversity part editing has some intersects with multimodal shape completion, here we use the metrics MMD, TMD, and UHD adopted by \citet{wu2020multimodal} to measure the quality, diversity, and fidelity of the edited results respectively. Our SGAS disentangles each part, so the input unedited parts can remain unchanged. Therefore, we only compare MMD and TMD. Considering previous part editing methods can only perform part editing on generated objects, which makes it impossible to obtain Gaussian noise corresponding to existing objects for editing. Hence, we use the representative multimodal shape completion method MSC-cGAN~\citep{wu2020multimodal} as the baseline for the this study. As shown in Table~\ref{tab:editings_metrics}, SGAS achieves excellent results in three representative categories.

\begin{table}[htb]
    \small
    \centering
    \caption{Ablation results for the hyperparameters $\alpha$ and $\beta$. The set of $\alpha:\beta$ can be determined by requirement.}
    \label{tab:ablation_alpha_beta}
    \begin{tabular}{c *6c}
        \toprule
        $\alpha:\beta$ & 1:10 & 1:2 & 1:1 & 2:1 & 5:1 & 10:1 \\
        \midrule
        MMD $\downarrow$ & 1.47 & 1.39 & \textbf{1.33} & 1.35 & 1.36 & 1.40 \\
        \midrule
        TMD $\uparrow$ & 1.89 & 2.65 & 4.36 & 4.75 & 5.34 & \textbf{7.68} \\
        \bottomrule
    \end{tabular}
\end{table}

During experiments, we find that the diversity metric TMD only measures the difference between edited results without considering fidelity and quality, which means two incorrect situations that can also result in high TMD: (a) the change of input unedited parts (corresponds to large UHD); (b) the edited parts with large differences but poor quality (corresponds to large MMD). Therefore, we further propose a new metric TMDS (TMD Surface) to solve the problems. Each point value on the surface is calculated as: 
\begin{equation}
\begin{gathered}
    {\rm TMDS}(\tau_{\rm uhd}, \tau_{\rm mmd}) = \mathop{mean}\limits_{p \in \mathbb{P}}
        \begin{cases}
            {\rm TMD}(s_1, .., s_k), \\
            \quad \text{if} \; \exists s_i, i=1...k, \\ 
            \quad \quad {\rm UHD}(p, s_i) \leq \tau_{\rm uhd} \\
            \quad \quad {\rm MMD}(s_i, \mathbb{D}) \leq \tau_{\rm mmd} \\
            0, \quad \text{otherwise} \\
        \end{cases} \\
    {\rm TMD}(s_1, .., s_k) = \sum_{i=1}^k \frac{1}{k-1} \sum_{j \neq i, j=1}^{k} {\rm CD}(s_i, s_j)
\end{gathered}
\end{equation}
where $p$ is input unedited parts, $s_i, i=1...k$ are K (e.g. 10) edited results. $\mathbb{P}$ is the unedited parts test dataset and $\mathbb{D}$ is orignal test dataset. CD is Chamfer Distance~\citep{Fan_2017_CVPR}. $\tau_{uhd}$ and $\tau_{mmd}$ are thresholds of UHD and MMD respectively. TMDS requires that each point value on the surface is calculated when K edited results are guaranteed to satisfy the corresponding UHD and MMD thresholds. The smaller the thresholds of MMD and UHD are, the more referential the calculated TMD is. Therefore, as shown in Figure~\ref{fig:tmds}, we can find that the editing diversity of SGAS is better than that of MSC-cGAN in the Chair category.

\begin{figure*}[htb]
    \centering
    \includegraphics[width=0.8\textwidth]{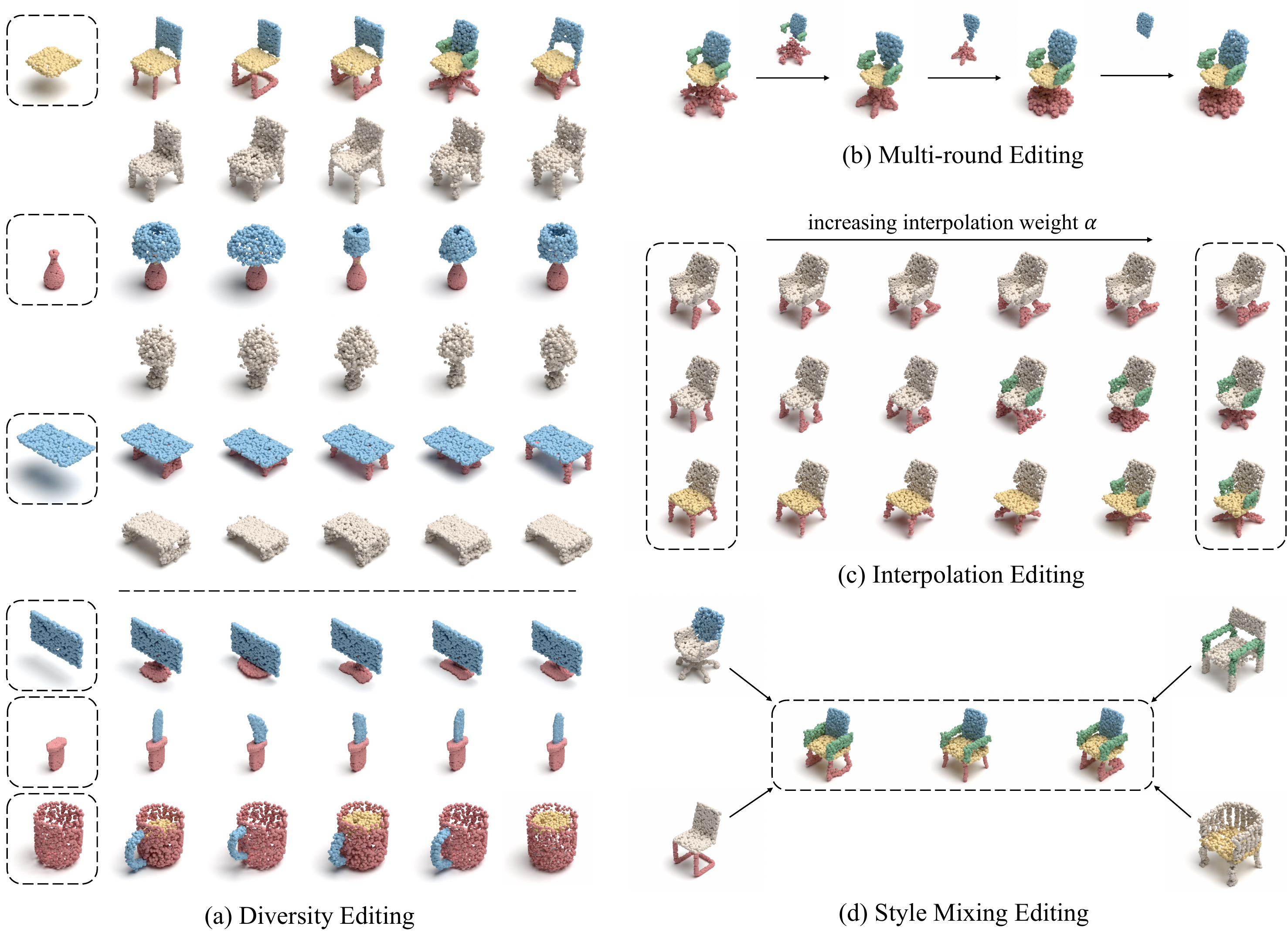}
    \caption{Various part editing operations. (a) Diversity editing comparison. The unedited parts are boxed, followed by five different edited results. The results of MSC-cGAN are uncolored while ours are colored by parts. (b) Re-editing of unsatisfactory edited results. The parts above the arrow are edited in each round. (c) Continuous transformation of selected parts (colored) through interpolation of two input Gaussian noises. Each row represents an interpolation of different number of parts. (d) Style Mixing of different parts in different objects. The mixed results are boxed.}
    \label{fig:editings}
\end{figure*}

We use SGAS to perform various part editing operations on PartNet.v0.Merged dataset. Figure~\ref{fig:editings}(a) is the visualized comparison of the diversity edited results. We also perform our SGAS on three new categories: Display, Knife, and Mug. It can be found that SGAS can not only keep the input unchanged but also have higher editing diversity and quality. Figure~\ref{fig:editings}(b) is a multi-round editing workflow achieved by SGAS. It demonstrates SGAS's ability to re-edit unsatisfactory edited results. Through three rounds of re-editing, a chair with right-angle arms and wheels is edited into a chair with circular arms and a circular base. Figure~\ref{fig:editings}(c) shows some interpolation part editings. If the chairs in the left and right box are generated by Gaussian noise $z_s$ and $z_t$ respectively, the chairs between boxes are generated by Gaussian noise $z = (1-\alpha)z_s + \alpha z_t$. $\alpha$ increases from 0 to 1. From top to bottom, each row represents an interpolation of one, two, and three parts. We can clearly observe the gradual change process of the parts. Figure~\ref{fig:editings}(d) demonstrates a style mixing part editing realized by SGAS. We first select four chairs with different styles. Then we edit different parts (colored) in different chairs. Finally, these edited parts are assembled to obtain chairs with styles from different chairs. This editing operation helps to create more diverse results.

\begin{figure}[htb]
    \centering
    \includegraphics[width=\columnwidth]{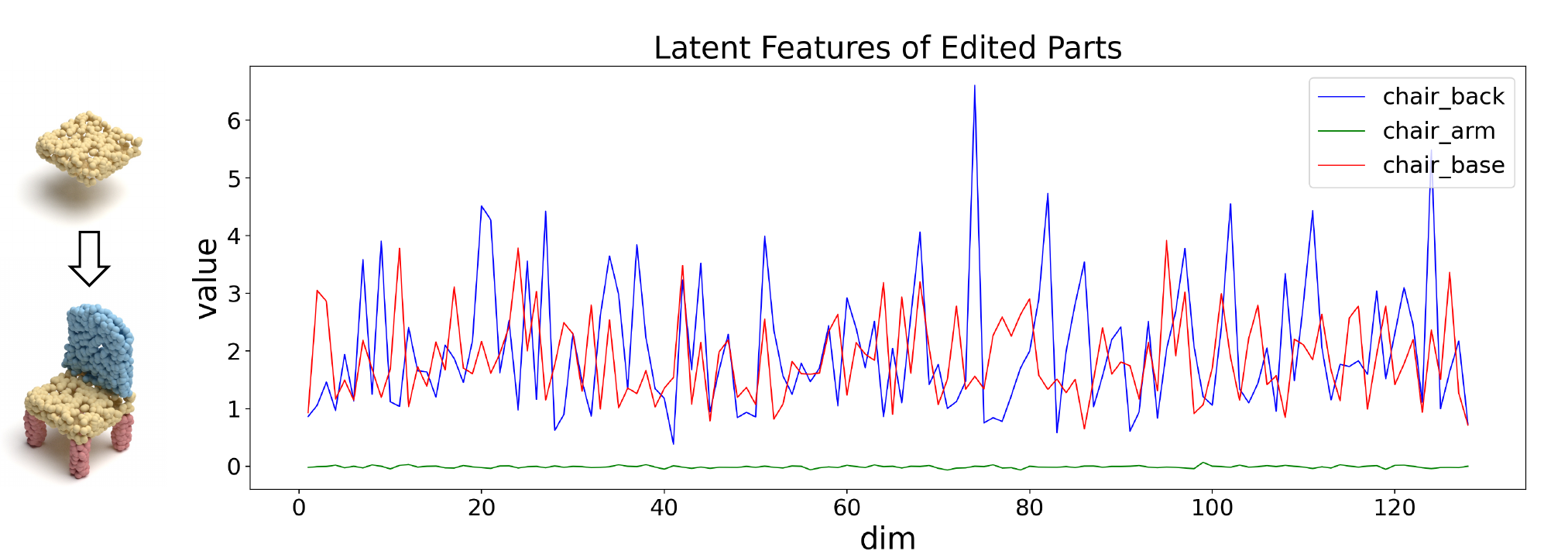}
    \caption{Edited parts with their corresponding latent features (128 dim). The chair without outputting arms has its corresponding latent feature of the chair arm near zero.}
    \label{fig:part-select}
\end{figure}

\begin{figure}[htb]
    \centering
    \includegraphics[width=\columnwidth]{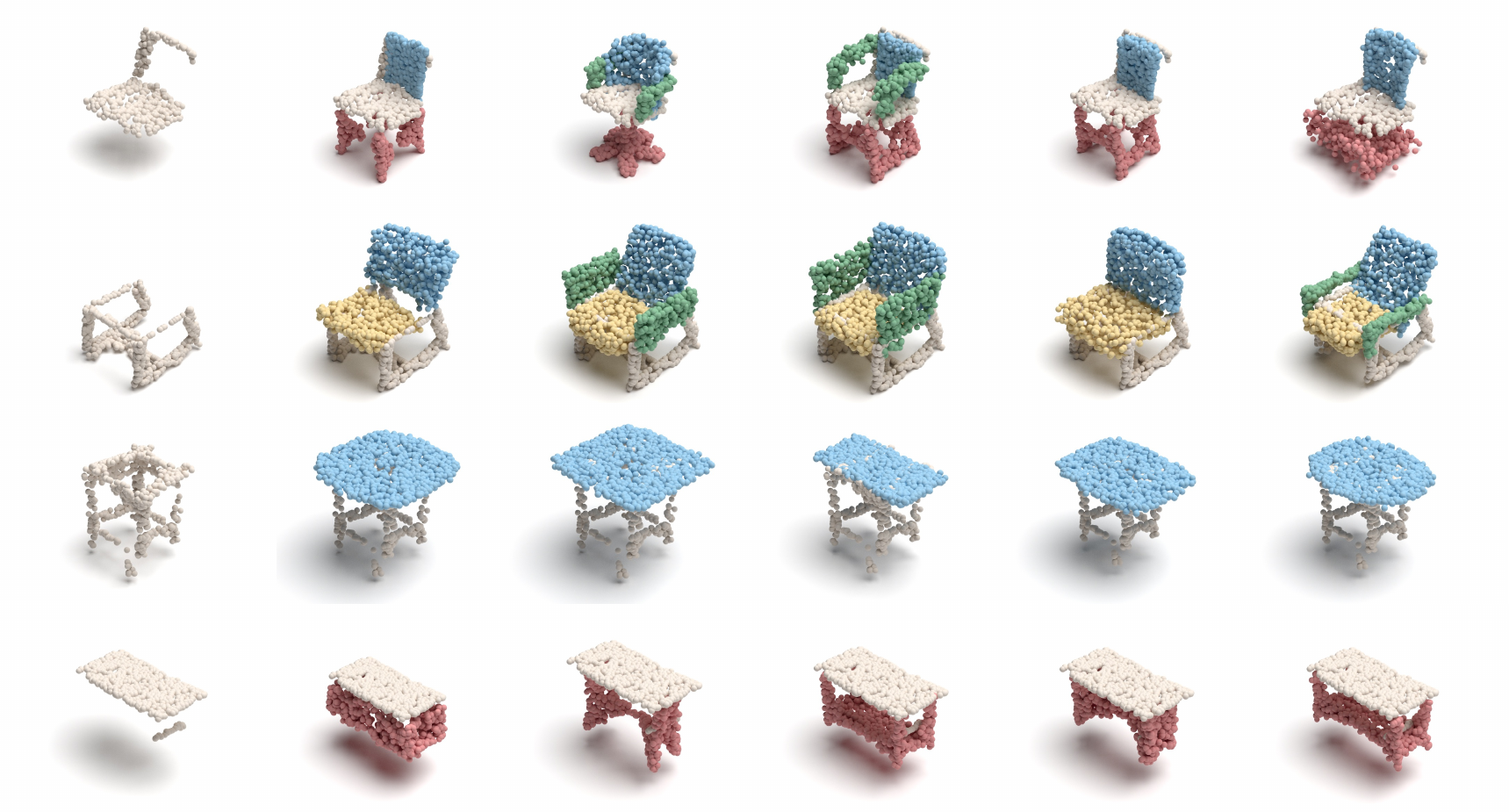}
    \caption{Performance on ScanNet. The leftmost column is incomplete input, followed by five diversity editing results.}
    \label{fig:real-scan}
\end{figure}

\begin{table}[htb]
    \small
    \centering
    \caption{Generative performance. The optimal and suboptimal results are highlighted in red and blue respectively. $M$ and $N$ represent the number of parts.}
    \label{tab:generation_metrics}
    \resizebox{\linewidth}{!}{\begin{tabular}{c c *5c}
        \toprule
        \multirow{2}{*}{Category} & \multirow{2}{*}{Model} & \multirow{2}{*}{JSD $\downarrow$} & \multicolumn{2}{c}{MMD $\downarrow$} & \multicolumn{2}{c}{COV \%, $\uparrow$} \\
        \cmidrule(lr){4-5} \cmidrule(lr){6-7}
        & & & CD & EMD & CD & EMD \\
        \midrule
        \multirow{5}{*}{Chair} & TreeGAN & 0.119 & \textcolor{blue}{0.0016} & 0.101 & 58 & 30 \\
        & MRGAN & 0.246 & 0.0021 & 0.166 & \textcolor{red}{67} & 23 \\
        & EditVAE (M=7) & 0.063 & \textcolor{red}{0.0014} & \textcolor{blue}{0.082} & 46 & 32 \\
        & EditVAE (M=3) & \textcolor{red}{0.031} & 0.0017 & 0.101 & 45 & \textcolor{blue}{39} \\
        & SGAS (N=7) & \textcolor{blue}{0.047} & 0.0020 & \textcolor{red}{0.076} & \textcolor{blue}{60} & \textcolor{red}{58} \\
        \midrule
        \multirow{5}{*}{Airplane} & TreeGAN & 0.097 & 0.0004 & 0.068 & 61 & 20 \\
        & MRGAN & 0.243 & 0.0006 & 0.114 & \textcolor{red}{75} & 21 \\
        & EditVAE (M=6) & \textcolor{blue}{0.043} & \textcolor{blue}{0.0004} & \textcolor{red}{0.024} & 39 & \textcolor{blue}{30} \\
        & EditVAE (M=3) & 0.044 & 0.0005 & 0.067 & 23 & 17 \\
        & SGAS (N=6) & \textcolor{red}{0.036} & \textcolor{red}{0.0004} & \textcolor{blue}{0.039} & \textcolor{blue}{61} & \textcolor{red}{58} \\
        \midrule
        \multirow{5}{*}{Table} & TreeGAN & 0.077 & 0.0018 & 0.082 & \textcolor{blue}{71} & \textcolor{blue}{48} \\
        & MRGAN & 0.287 & 0.0020 & 0.155 & \textcolor{red}{78} & 31 \\
        & EditVAE (M=5) & 0.081 & \textcolor{red}{0.0016} & \textcolor{blue}{0.071} & 42 & 27 \\
        & EditVAE (M=3) & \textcolor{red}{0.042} & \textcolor{blue}{0.0017} & 0.130 & 39 & 30 \\
        & SGAS (N=5) & \textcolor{blue}{0.057} & 0.0020 & \textcolor{red}{0.069} & 65 & \textcolor{red}{65} \\
        \bottomrule
    \end{tabular}}
\end{table}

\begin{figure}[htb]
    \centering
    \includegraphics[width=0.7\columnwidth]{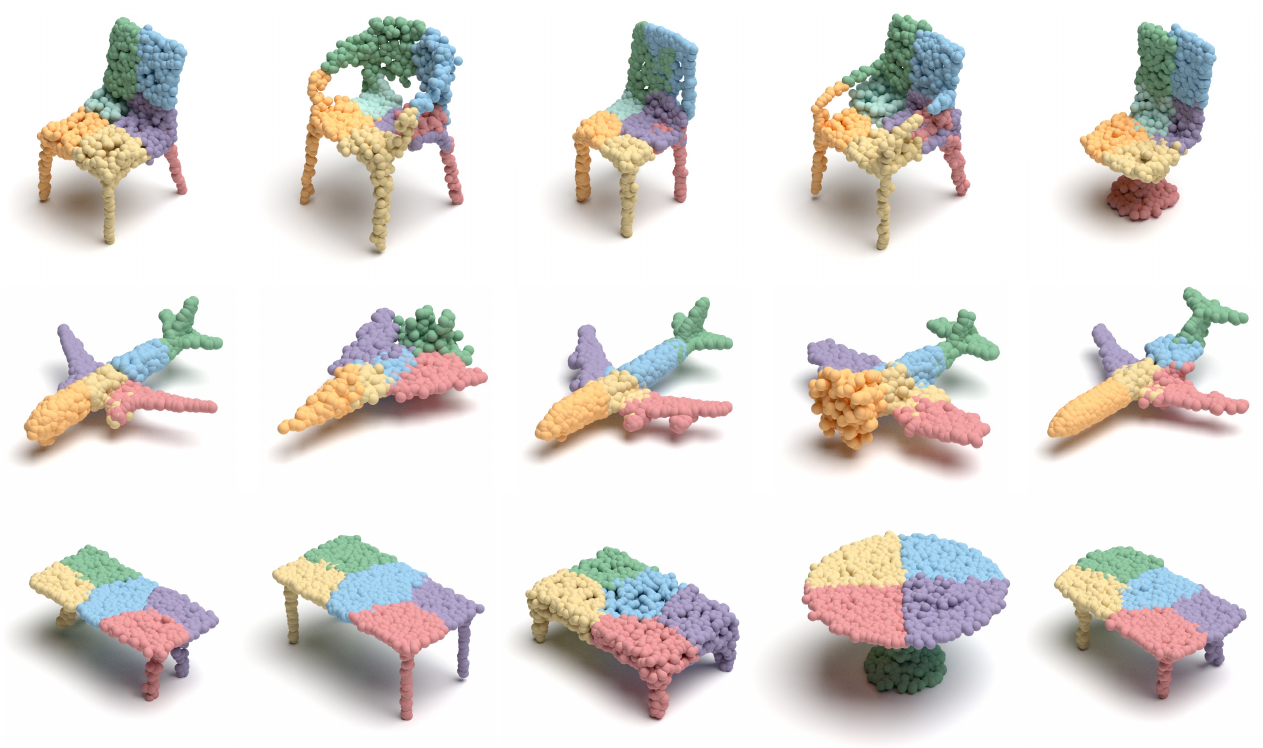}
    \caption{Point clouds generated by SGAS, colored by parts.}
    \label{fig:generation}
\end{figure}

As the hyperparameters $\alpha$ and $\beta$ represent diversity and quality respectively, and have a significant impact on the edited results. Hence, we conduct an ablation study on them. The results are presented in Table~\ref{tab:ablation_alpha_beta}. It is observed that as $\alpha:\beta$ increase, the TMD also increase. However, the MMD, which measures the quality of the edited results, initially improves but then worsens. Therefore, we finally choose $\alpha:\beta=1:1$ to realize part editing. However, if diversity is more important for the edited results, a higher $\alpha:\beta$ can also be used. To demonstrate SGAS's ability to automatically select parts, we visualize the latent features of the edited parts. As shown in Figure~\ref{fig:part-select}, it can be found that the latent feature of the chair arm is near zero, which means SGAS believes that the newly generated chair arm is inappropriate. Therefore, the Part Select module in SGAS will filter this latent feature to prevent the chair arm from being output. To further prove the generalization ability of SGAS, we train SGAS on PartNet.v0.Merged dataset and test it on ScanNet~\citep{dai2017scannet} dataset. The results can be found in Figure~\ref{fig:real-scan}. For parts with high missing rates, we will regenerate them, such as the chair back in the 1st row, and the newly generated parts are compatible with the existing incomplete parts. For parts with low missing rates, we will keep them directly. It can be found that even on unseen objects, SGAS's diversity editing results are still good.

\subsection{Unsupervised Part-aware Point Cloud Generation}

By pruning SGAS, including removing unedited parts input, AdaIN Layer, Part Mask, and Part Select, SGAS can be applied to realize unsupervised part-aware point cloud generation. It includes two steps: (a) modifying an unsupervised shape co-segmentation method AXform~\citep{Zhang_Yang_Wu_Jin_2022} to get Ground Truth part datasets; (b) training SGAS on these part datasets. Specifically, we first modify the multi-branch AXform to output one structure point per branch. Second, we use these structure points to co-segment the Ground Truth point clouds into $n$ part datasets. Third, the segmented parts are pre-encoded into latent features. Finally, these latent features are used to supervise the training of SGAS. However, we found that there might be some large gaps between parts during generation. Therefore, to achieve seamless generation, during unsupervised part segmentation, we further expand the number of points per part from $\left\lfloor\frac{2048}{n}\right\rfloor$ to $(1+\gamma)\left\lfloor\frac{2048}{n}\right\rfloor$. Here $\gamma = 0.1$. In the final output, we downsample the point cloud to 2048 points.

\begin{table}[htb]
    \small
    \centering
    \caption{Ablation results for the number of parts. The optimal and suboptimal results are highlighted in red and blue respectively. $n=6$ is a suitable number of parts.}
    \label{tab:ablation_part_num}
    \begin{tabular}{c *5c}
        \toprule
        \multirow{2}{*}{\#Parts (n)} & \multirow{2}{*}{JSD $\downarrow$} & \multicolumn{2}{c}{MMD $\downarrow$} & \multicolumn{2}{c}{COV \%,$\uparrow$} \\
        \cmidrule(lr){3-4} \cmidrule(lr){5-6}
        & & CD & EMD & CD & EMD \\
        \midrule
        2  & 0.042 & 0.0004 & 0.045 & \textcolor{blue}{61} & 47 \\
        3  & 0.039 & \textcolor{blue}{0.0004} & 0.043	& 60 & 52 \\
        5  & 0.040 & 0.0005 & 0.042	& 60 & 50 \\
        6  & \textcolor{blue}{0.036} & \textcolor{red}{0.0004} & \textcolor{red}{0.039} & \textcolor{red}{61} & \textcolor{red}{58} \\
        8  & 0.040 & 0.0005 & 0.044	& 60 & 46 \\
        13 & \textcolor{red}{0.035} & 0.0005 & \textcolor{blue}{0.040} & 60 & \textcolor{blue}{53} \\
        \bottomrule
    \end{tabular}
\end{table}

The quantitative results are shown in Table~\ref{tab:generation_metrics}. The metrics are proposed by \citet{achlioptas2017latent_pc}, and the results of previous methods are obtained from EditVAE~\citep{Li_Liu_Walder_2022}. MMD and COV represent the quality and diversity of the generated results respectively. It can be found that SGAS achieves excellent results overall, with the most number of top two metrics. Especially on COV-EMD, which represents diversity, SGAS has a significant improvement. Figure~\ref{fig:generation} gives visualized results of the generated point clouds. Different colors correspond to different parts. It intuitively illustrates the diversity and quality of the results generated by SGAS. We also conduct an ablation study on the number of parts in the Airplane category. As shown in Table~\ref{tab:ablation_part_num}, more or fewer parts are not necessarily beneficial to the results. Therefore, we chose the number of parts $N=6$.

\section{Conclusion}

Previous methods do not disentangle each part completely or lack constraints between parts, which leads to poor diversity, fidelity, and quality when performing part editing. In this work, to solve these problems, we first propose a novel four-stage point cloud part editing process. Then based on this process and two new strategies: feature disentanglement and constraint, we propose a part editing model SGAS. It can realize various part editing operations. By introducing metrics from multimodal completion and proposing a new metric TMDS, we establish quantitative evaluations for diversity part editing. In addition, SGAS can be pruned to realize unsupervised part-aware point cloud generation. Experiments show that it performs better than previous methods.

\noindent \textbf{Limitation} Since we do not design part offset networks for the generated parts but instead utilize the relatively fixed spatial positions of each generated part to ensure good assembly, SGAS can only achieve part editing for objects with relatively consistent prototypes. For example, SGAS cannot handle part editing for both ceiling lamps and table lamps simultaneously as the spatial order of the parts is opposite. In addition, we also find that the performance of SGAS is limited by the pre-trained autoencoders. The embedded features are better when the parts are normalized. Therefore, it will be beneficial to first generate normalized parts and then design part offset networks to align them in the future.

\section{Acknowledgments}

This work was supported by National Natural Science Fund of China (62176064). Cheng Jin is the corresponding author.

\bibliography{aaai24}

\begin{thebibliography}{49}
\providecommand{\natexlab}[1]{#1}

\bibitem[{Achlioptas et~al.(2017)Achlioptas, Diamanti, Mitliagkas, and
  Guibas}]{achlioptas2017latent_pc}
Achlioptas, P.; Diamanti, O.; Mitliagkas, I.; and Guibas, L.~J. 2017.
\newblock Learning Representations and Generative Models For 3D Point Clouds.
\newblock \emph{arXiv preprint arXiv:1707.02392}.

\bibitem[{Arjovsky, Chintala, and Bottou(2017)}]{pmlr-v70-arjovsky17a}
Arjovsky, M.; Chintala, S.; and Bottou, L. 2017.
\newblock {W}asserstein Generative Adversarial Networks.
\newblock In \emph{Proceedings of the 34th International Conference on Machine
  Learning}, volume~70, 214--223.

\bibitem[{Arora et~al.(2022)Arora, Mishra, Peng, Li, and
  Mahdavi-Amiri}]{Arora_2022_CVPR}
Arora, H.; Mishra, S.; Peng, S.; Li, K.; and Mahdavi-Amiri, A. 2022.
\newblock Multimodal Shape Completion via Implicit Maximum Likelihood
  Estimation.
\newblock In \emph{Proceedings of the IEEE/CVF Conference on Computer Vision
  and Pattern Recognition (CVPR) Workshops}, 2958--2967.

\bibitem[{Chen, Chen, and Mitra(2020)}]{chen2020pcl2pcl}
Chen, X.; Chen, B.; and Mitra, N.~J. 2020.
\newblock Unpaired Point Cloud Completion on Real Scans using Adversarial
  Training.
\newblock In \emph{Proceedings of the International Conference on Learning
  Representations (ICLR)}.

\bibitem[{Chen et~al.(2019)Chen, Yin, Fisher, Chaudhuri, and
  Zhang}]{Chen_2019_ICCV}
Chen, Z.; Yin, K.; Fisher, M.; Chaudhuri, S.; and Zhang, H. 2019.
\newblock BAE-NET: Branched Autoencoder for Shape Co-Segmentation.
\newblock In \emph{Proceedings of the IEEE/CVF International Conference on
  Computer Vision (ICCV)}.

\bibitem[{Cheng et~al.(2022)Cheng, Li, Liu, Sun, and
  Yang}]{cheng2022autoregressive}
Cheng, A.-C.; Li, X.; Liu, S.; Sun, M.; and Yang, M.-H. 2022.
\newblock Autoregressive 3d shape generation via canonical mapping.
\newblock In \emph{Computer Vision--ECCV 2022: 17th European Conference, Tel
  Aviv, Israel, October 23--27, 2022, Proceedings, Part III}, 89--104.

\bibitem[{Cheng et~al.(2023)Cheng, Lee, Tuyakov, Schwing, and
  Gui}]{Cheng_2023_CVPR}
Cheng, Y.-C.; Lee, H.-Y.; Tuyakov, S.; Schwing, A.; and Gui, L. 2023.
\newblock {SDFusion}: Multimodal 3D Shape Completion, Reconstruction, and
  Generation.
\newblock In \emph{Proceedings of the IEEE/CVF Conference on Computer Vision
  and Pattern Recognition (CVPR)}.

\bibitem[{Dai et~al.(2017)Dai, Chang, Savva, Halber, Funkhouser, and
  Nie{\ss}ner}]{dai2017scannet}
Dai, A.; Chang, A.~X.; Savva, M.; Halber, M.; Funkhouser, T.; and Nie{\ss}ner,
  M. 2017.
\newblock ScanNet: Richly-annotated 3D Reconstructions of Indoor Scenes.
\newblock In \emph{Proc. Computer Vision and Pattern Recognition (CVPR), IEEE}.

\bibitem[{Dubrovina et~al.(2019)Dubrovina, Xia, Achlioptas, Shalah, Groscot,
  and Guibas}]{Dubrovina_2019_ICCV}
Dubrovina, A.; Xia, F.; Achlioptas, P.; Shalah, M.; Groscot, R.; and Guibas,
  L.~J. 2019.
\newblock Composite Shape Modeling via Latent Space Factorization.
\newblock In \emph{Proceedings of the IEEE/CVF International Conference on
  Computer Vision (ICCV)}.

\bibitem[{Fan, Su, and Guibas(2017)}]{Fan_2017_CVPR}
Fan, H.; Su, H.; and Guibas, L.~J. 2017.
\newblock A Point Set Generation Network for 3D Object Reconstruction From a
  Single Image.
\newblock In \emph{Proceedings of the IEEE Conference on Computer Vision and
  Pattern Recognition (CVPR)}.

\bibitem[{Gal et~al.(2021)Gal, Bermano, Zhang, and Cohen-Or}]{Gal_2021_ICCV}
Gal, R.; Bermano, A.; Zhang, H.; and Cohen-Or, D. 2021.
\newblock MRGAN: Multi-Rooted 3D Shape Representation Learning With
  Unsupervised Part Disentanglement.
\newblock In \emph{Proceedings of the IEEE/CVF International Conference on
  Computer Vision (ICCV) Workshops}, 2039--2048.

\bibitem[{Gulrajani et~al.(2017)Gulrajani, Ahmed, Arjovsky, Dumoulin, and
  Courville}]{NIPS2017_892c3b1c}
Gulrajani, I.; Ahmed, F.; Arjovsky, M.; Dumoulin, V.; and Courville, A.~C.
  2017.
\newblock Improved Training of Wasserstein GANs.
\newblock In \emph{Advances in Neural Information Processing Systems},
  volume~30.

\bibitem[{Huang and Belongie(2017)}]{Huang_2017_ICCV}
Huang, X.; and Belongie, S. 2017.
\newblock Arbitrary Style Transfer in Real-Time With Adaptive Instance
  Normalization.
\newblock In \emph{Proceedings of the IEEE International Conference on Computer
  Vision (ICCV)}.

\bibitem[{Hui et~al.(2022)Hui, Li, Hu, and Fu}]{Hui_2022_CVPR}
Hui, K.-H.; Li, R.; Hu, J.; and Fu, C.-W. 2022.
\newblock Neural Template: Topology-Aware Reconstruction and Disentangled
  Generation of 3D Meshes.
\newblock In \emph{Proceedings of the IEEE/CVF Conference on Computer Vision
  and Pattern Recognition (CVPR)}, 18572--18582.

\bibitem[{Jiang and Daniilidis(2022)}]{jiang2022probabilistic}
Jiang, W.; and Daniilidis, K. 2022.
\newblock Probabilistic Shape Completion by Estimating Canonical Factors with
  Hierarchical VAE.
\newblock \emph{arXiv preprint arXiv:2212.03370}.

\bibitem[{Jones et~al.(2020)Jones, Barton, Xu, Wang, Jiang, Guerrero, Mitra,
  and Ritchie}]{jones2020shapeAssembly}
Jones, R.~K.; Barton, T.; Xu, X.; Wang, K.; Jiang, E.; Guerrero, P.; Mitra,
  N.~J.; and Ritchie, D. 2020.
\newblock ShapeAssembly: Learning to Generate Programs for 3D Shape Structure
  Synthesis.
\newblock \emph{ACM Transactions on Graphics (TOG), Siggraph Asia 2020}, 39(6):
  Article 234.

\bibitem[{Li, Niu, and Xu(2020)}]{Li_Niu_Xu_2020}
Li, J.; Niu, C.; and Xu, K. 2020.
\newblock Learning Part Generation and Assembly for Structure-Aware Shape
  Synthesis.
\newblock \emph{Proceedings of the AAAI Conference on Artificial Intelligence},
  34(07): 11362--11369.

\bibitem[{Li et~al.(2017)Li, Xu, Chaudhuri, Yumer, Zhang, and
  Guibas}]{li_sig17}
Li, J.; Xu, K.; Chaudhuri, S.; Yumer, E.; Zhang, H.; and Guibas, L. 2017.
\newblock GRASS: Generative Recursive Autoencoders for Shape Structures.
\newblock \emph{ACM Transactions on Graphics (Proc. of SIGGRAPH 2017)}, 36(4).

\bibitem[{Li et~al.(2021)Li, Li, Hui, and Fu}]{li2021spgan}
Li, R.; Li, X.; Hui, K.-H.; and Fu, C.-W. 2021.
\newblock {SP-GAN}:Sphere-Guided 3D Shape Generation and Manipulation.
\newblock \emph{ACM Transactions on Graphics (Proc. SIGGRAPH)}, 40(4).

\bibitem[{Li, Liu, and Walder(2022)}]{Li_Liu_Walder_2022}
Li, S.; Liu, M.; and Walder, C. 2022.
\newblock EditVAE: Unsupervised Parts-Aware Controllable 3D Point Cloud Shape
  Generation.
\newblock \emph{Proceedings of the AAAI Conference on Artificial Intelligence},
  36(2): 1386--1394.

\bibitem[{Li and Baciu(2022)}]{Li_2022_TVCG}
Li, Y.; and Baciu, G. 2022.
\newblock SG-GAN: Adversarial Self-Attention GCN for Point Cloud Topological
  Parts Generation.
\newblock \emph{IEEE Transactions on Visualization and Computer Graphics},
  28(10): 3499--3512.

\bibitem[{Liu et~al.(2021)Liu, Snodgrass, Khalifa, Risi, Yannakakis, and
  Togelius}]{Liu_2021_NCA}
Liu, J.; Snodgrass, S.; Khalifa, A.; Risi, S.; Yannakakis, G.~N.; and Togelius,
  J. 2021.
\newblock Deep learning for procedural content generation.
\newblock \emph{Neural Computing and Applications}, 33(1): 19--37.

\bibitem[{Mittal et~al.(2022)Mittal, Cheng, Singh, and
  Tulsiani}]{Mittal_2022_CVPR}
Mittal, P.; Cheng, Y.-C.; Singh, M.; and Tulsiani, S. 2022.
\newblock AutoSDF: Shape Priors for 3D Completion, Reconstruction and
  Generation.
\newblock In \emph{Proceedings of the IEEE/CVF Conference on Computer Vision
  and Pattern Recognition (CVPR)}, 306--315.

\bibitem[{Mo et~al.(2019{\natexlab{a}})Mo, Guerrero, Yi, Su, Wonka, Mitra, and
  Guibas}]{mo2019structurenet}
Mo, K.; Guerrero, P.; Yi, L.; Su, H.; Wonka, P.; Mitra, N.; and Guibas, L.
  2019{\natexlab{a}}.
\newblock StructureNet: Hierarchical Graph Networks for 3D Shape Generation.
\newblock \emph{ACM Transactions on Graphics (TOG), Siggraph Asia 2019}, 38(6):
  Article 242.

\bibitem[{Mo et~al.(2020)Mo, Guerrero, Yi, Su, Wonka, Mitra, and
  Guibas}]{Mo_2020_CVPR}
Mo, K.; Guerrero, P.; Yi, L.; Su, H.; Wonka, P.; Mitra, N.~J.; and Guibas,
  L.~J. 2020.
\newblock StructEdit: Learning Structural Shape Variations.
\newblock In \emph{Proceedings of the IEEE/CVF Conference on Computer Vision
  and Pattern Recognition (CVPR)}.

\bibitem[{Mo et~al.(2019{\natexlab{b}})Mo, Zhu, Chang, Yi, Tripathi, Guibas,
  and Su}]{Mo_2019_CVPR}
Mo, K.; Zhu, S.; Chang, A.~X.; Yi, L.; Tripathi, S.; Guibas, L.~J.; and Su, H.
  2019{\natexlab{b}}.
\newblock {PartNet}: A Large-Scale Benchmark for Fine-Grained and Hierarchical
  Part-Level {3D} Object Understanding.
\newblock In \emph{The IEEE Conference on Computer Vision and Pattern
  Recognition (CVPR)}.

\bibitem[{{\"O}ng{\"u}n and Temizel(2020)}]{ngn2020LPMNetLP}
{\"O}ng{\"u}n, C.; and Temizel, A. 2020.
\newblock LPMNet: Latent part modification and generation for 3D point clouds.
\newblock \emph{Comput. Graph.}, 96: 1--13.

\bibitem[{Postels et~al.(2021)Postels, Liu, Spezialetti, Gool, and
  Tombari}]{Postels_2021_3DV}
Postels, J.; Liu, M.; Spezialetti, R.; Gool, L.~V.; and Tombari, F. 2021.
\newblock Go with the Flows: Mixtures of Normalizing Flows for Point Cloud
  Generation and Reconstruction.
\newblock In \emph{2021 International Conference on 3D Vision (3DV)},
  1249--1258.

\bibitem[{Qi et~al.(2017)Qi, Su, Mo, and Guibas}]{Qi_2017_CVPR}
Qi, C.~R.; Su, H.; Mo, K.; and Guibas, L.~J. 2017.
\newblock PointNet: Deep Learning on Point Sets for 3D Classification and
  Segmentation.
\newblock In \emph{Proceedings of the IEEE Conference on Computer Vision and
  Pattern Recognition (CVPR)}.

\bibitem[{Schor et~al.(2019)Schor, Katzir, Zhang, and
  Cohen-Or}]{Schor_2019_ICCV}
Schor, N.; Katzir, O.; Zhang, H.; and Cohen-Or, D. 2019.
\newblock CompoNet: Learning to Generate the Unseen by Part Synthesis and
  Composition.
\newblock In \emph{The IEEE International Conference on Computer Vision
  (ICCV)}.

\bibitem[{Shu, Park, and Kwon(2019)}]{Shu_2019_ICCV}
Shu, D.~W.; Park, S.~W.; and Kwon, J. 2019.
\newblock 3D Point Cloud Generative Adversarial Network Based on Tree
  Structured Graph Convolutions.
\newblock In \emph{Proceedings of the IEEE/CVF International Conference on
  Computer Vision (ICCV)}.

\bibitem[{van~den Oord, Vinyals, and kavukcuoglu(2017)}]{NIPS2017_7a98af17}
van~den Oord, A.; Vinyals, O.; and kavukcuoglu, k. 2017.
\newblock Neural Discrete Representation Learning.
\newblock In \emph{Advances in Neural Information Processing Systems},
  volume~30.

\bibitem[{Wang et~al.(2018)Wang, Schor, Hu, Huang, Cohen-Or, and
  Huang}]{Wang_TOG_2018}
Wang, H.; Schor, N.; Hu, R.; Huang, H.; Cohen-Or, D.; and Huang, H. 2018.
\newblock Global-to-Local Generative Model for 3D Shapes.
\newblock \emph{ACM Trans. Graph.}, 37(6).

\bibitem[{Wang et~al.(2022)Wang, Guerrero, Kim, Chaudhuri, Sung, and
  Ritchie}]{wang2022shape}
Wang, K.; Guerrero, P.; Kim, V.~G.; Chaudhuri, S.; Sung, M.; and Ritchie, D.
  2022.
\newblock The shape part slot machine: Contact-based reasoning for generating
  3D shapes from parts.
\newblock In \emph{Computer Vision--ECCV 2022: 17th European Conference, Tel
  Aviv, Israel, October 23--27, 2022, Proceedings, Part III}, 610--626.

\bibitem[{Wu et~al.(2023)Wu, Zheng, Pfrommer, and Beyerer}]{wu2023attention}
Wu, C.; Zheng, J.; Pfrommer, J.; and Beyerer, J. 2023.
\newblock Attention-based Part Assembly for 3D Volumetric Shape Modeling.
\newblock \emph{arXiv preprint arXiv:2304.10986}.

\bibitem[{Wu et~al.(2020{\natexlab{a}})Wu, Chen, Zhuang, and
  Chen}]{wu2020multimodal}
Wu, R.; Chen, X.; Zhuang, Y.; and Chen, B. 2020{\natexlab{a}}.
\newblock Multimodal shape completion via conditional generative adversarial
  networks.
\newblock In \emph{Computer Vision--ECCV 2020: 16th European Conference,
  Glasgow, UK, August 23--28, 2020, Proceedings, Part IV 16}, 281--296.

\bibitem[{Wu et~al.(2020{\natexlab{b}})Wu, Zhuang, Xu, Zhang, and
  Chen}]{Wu_2020_CVPR}
Wu, R.; Zhuang, Y.; Xu, K.; Zhang, H.; and Chen, B. 2020{\natexlab{b}}.
\newblock PQ-NET: A Generative Part Seq2Seq Network for 3D Shapes.
\newblock In \emph{Proceedings of the IEEE/CVF Conference on Computer Vision
  and Pattern Recognition (CVPR)}.

\bibitem[{Wu et~al.(2019)Wu, Wang, Lin, Lischinski, Cohen-Or, and
  Huang}]{Wu_TOG_2019}
Wu, Z.; Wang, X.; Lin, D.; Lischinski, D.; Cohen-Or, D.; and Huang, H. 2019.
\newblock SAGNet: Structure-Aware Generative Network for 3D-Shape Modeling.
\newblock \emph{ACM Trans. Graph.}, 38(4).

\bibitem[{Yan et~al.(2022)Yan, Lin, Mitra, Lischinski, Cohen-Or, and
  Huang}]{Yan_2022_CVPR}
Yan, X.; Lin, L.; Mitra, N.~J.; Lischinski, D.; Cohen-Or, D.; and Huang, H.
  2022.
\newblock ShapeFormer: Transformer-Based Shape Completion via Sparse
  Representation.
\newblock In \emph{Proceedings of the IEEE/CVF Conference on Computer Vision
  and Pattern Recognition (CVPR)}, 6239--6249.

\bibitem[{Yang et~al.(2018)Yang, Feng, Shen, and Tian}]{Yang_2018_CVPR}
Yang, Y.; Feng, C.; Shen, Y.; and Tian, D. 2018.
\newblock FoldingNet: Point Cloud Auto-Encoder via Deep Grid Deformation.
\newblock In \emph{Proceedings of the IEEE Conference on Computer Vision and
  Pattern Recognition (CVPR)}.

\bibitem[{Yi et~al.(2016)Yi, Kim, Ceylan, Shen, Yan, Su, Lu, Huang, Sheffer,
  and Guibas}]{Yi16}
Yi, L.; Kim, V.~G.; Ceylan, D.; Shen, I.-C.; Yan, M.; Su, H.; Lu, C.; Huang,
  Q.; Sheffer, A.; and Guibas, L. 2016.
\newblock A Scalable Active Framework for Region Annotation in 3D Shape
  Collections.
\newblock \emph{ACM Trans. Graph.}, 35(6).

\bibitem[{Yin et~al.(2020)Yin, Chen, Chaudhuri, Fisher, Kim, and
  Zhang}]{yin2020coalesce}
Yin, K.; Chen, Z.; Chaudhuri, S.; Fisher, M.; Kim, V.~G.; and Zhang, H. 2020.
\newblock Coalesce: Component assembly by learning to synthesize connections.
\newblock In \emph{2020 International Conference on 3D Vision (3DV)}, 61--70.

\bibitem[{Zhang et~al.(2021)Zhang, Choi, Kim, and Kim}]{zhang2021gca}
Zhang, D.; Choi, C.; Kim, J.; and Kim, Y.~M. 2021.
\newblock Learning to Generate 3D Shapes with Generative Cellular Automata.
\newblock In \emph{International Conference on Learning Representations}.

\bibitem[{Zhang et~al.(2022)Zhang, Yang, Wu, and Jin}]{Zhang_Yang_Wu_Jin_2022}
Zhang, K.; Yang, X.; Wu, Y.; and Jin, C. 2022.
\newblock Attention-Based Transformation from Latent Features to Point Clouds.
\newblock \emph{Proceedings of the AAAI Conference on Artificial Intelligence},
  36(3): 3291--3299.

\bibitem[{Zhao et~al.(2021)Zhao, Zhou, Chen, Hu, and Ai}]{Zhao_2021_MM}
Zhao, Y.; Zhou, Y.; Chen, R.; Hu, B.; and Ai, X. 2021.
\newblock MM-Flow: Multi-Modal Flow Network for Point Cloud Completion.
\newblock In \emph{Proceedings of the 29th ACM International Conference on
  Multimedia}, 3266–3274.

\bibitem[{Zhou, Du, and Wu(2021)}]{Zhou_2021_ICCV}
Zhou, L.; Du, Y.; and Wu, J. 2021.
\newblock 3D Shape Generation and Completion Through Point-Voxel Diffusion.
\newblock In \emph{Proceedings of the IEEE/CVF International Conference on
  Computer Vision (ICCV)}, 5826--5835.

\bibitem[{Zhu et~al.(2020)Zhu, Xu, Chaudhuri, Yi, Guibas, and
  Zhang}]{Zhu_2020_CVPR}
Zhu, C.; Xu, K.; Chaudhuri, S.; Yi, L.; Guibas, L.~J.; and Zhang, H. 2020.
\newblock AdaCoSeg: Adaptive Shape Co-Segmentation With Group Consistency Loss.
\newblock In \emph{Proceedings of the IEEE/CVF Conference on Computer Vision
  and Pattern Recognition (CVPR)}.

\bibitem[{Zhuang(2022)}]{Zhuang_2022}
Zhuang, Y. 2022.
\newblock Progressive Multimodal Shape Generation via Contextual Part
  Reasoning.
\newblock In \emph{2022 The 6th International Conference on Machine Learning
  and Soft Computing}, 173–178.

\bibitem[{Zou et~al.(2017)Zou, Yumer, Yang, Ceylan, and Hoiem}]{Zou_2017_ICCV}
Zou, C.; Yumer, E.; Yang, J.; Ceylan, D.; and Hoiem, D. 2017.
\newblock 3D-PRNN: Generating Shape Primitives With Recurrent Neural Networks.
\newblock In \emph{Proceedings of the IEEE International Conference on Computer
  Vision (ICCV)}.

\end{thebibliography}

\end{document}